\documentclass[11pt]{article}

\usepackage[final]{acl}

\usepackage{times}
\usepackage{latexsym}

\usepackage[T1]{fontenc}

\usepackage[utf8]{inputenc}

\usepackage{microtype}

\usepackage{inconsolata}

\usepackage{graphicx}

\usepackage{booktabs}

\usepackage{amsmath}

\usepackage{tikz}
\usetikzlibrary{arrows.meta,positioning,shapes.geometric,fit,backgrounds}
\usepackage{pgfplots}
\pgfplotsset{compat=1.18}
\definecolor{cLLM}{HTML}{2C6FBB}
\definecolor{cRegex}{HTML}{D1892F}
\definecolor{cEmbed}{HTML}{5A9E6F}
\definecolor{cInk}{HTML}{333333}
\definecolor{cBox}{HTML}{F2F4F7}

\title{Topic Matching in the Wild: Benchmark and Lessons from Real-World ASR Transcripts}

\author{
  Saman Rahbar,  Xiliang Zhu,  Irvin Cardoza,  David Rossouw \\
  Dialpad Inc. \\
  \texttt{\{sam.rahbar, xzhu, irvin.cardoza, davidr\}@dialpad.com}
}

\begin{document}
\maketitle

\begin{abstract}
In contact centers, real-time agent-assist tools determine, for each of many predefined topics, whether a live customer utterance is relevant and display a coaching card to the agent when it is. The input is noisy and challenging: ASR (Automatic Speech Recognition) transcripts of spontaneous phone conversations, which can be unclear, repetitive, and mostly lack punctuation. To systematically study this real-world task, we curate a human-annotated topic-utterance judgments dataset sourced from real call-center transcripts. We compare three types of matchers: a regex-based baseline, zero-shot sentence-embedding encoders, and Gemini-based LLM matchers. In addition, two types of topic representations are studied in our benchmark: keyphrases and natural language description. Our empirical experiments highlight the superior performance of lightweight LLM matchers over embedding and regex models when equipped with natural language descriptions.

\end{abstract}

\section{Introduction}

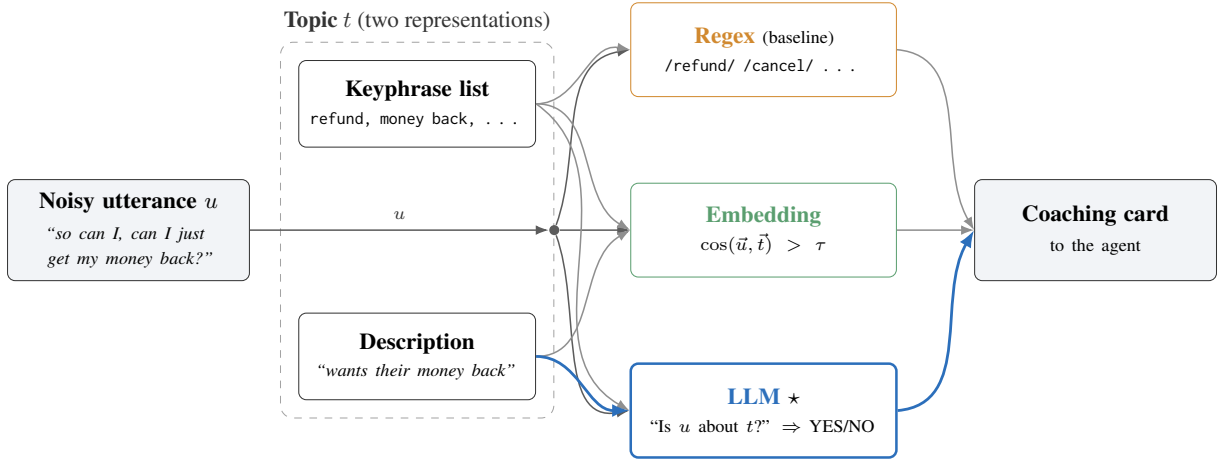
\begin{figure*}[t]
\centering
\resizebox{\textwidth}{!}{%
\begin{tikzpicture}[
  font=\small,
  io/.style   ={draw=cInk, rounded corners=3pt, fill=cBox, align=center, inner sep=5pt, text width=30mm, minimum height=14mm},
  rep/.style  ={draw=cInk, rounded corners=3pt, fill=white, align=center, inner sep=4pt, text width=30mm, minimum height=12mm},
  mat/.style  ={draw=cInk, rounded corners=3pt, fill=white, align=center, inner sep=4pt, text width=34mm, minimum height=13mm},
  edge/.style ={draw=cInk!55, -{Latex[length=1.8mm]}, semithick},
  ustream/.style={draw=cInk!80, -{Latex[length=1.8mm]}, semithick},
  best/.style ={draw=cLLM, -{Latex[length=2.4mm]}, line width=1.1pt},
  jdot/.style ={circle, fill=cInk!80, inner sep=1.4pt},
]
\node[io] (u) at (-6.3,0)
  {\textbf{Noisy utterance}~$u$\\[2pt]{\scriptsize\itshape ``so can I, can I just get my money back?''}};

\node[rep] (kp)   at (-2.3, 1.75)
  {\textbf{Keyphrase list}\\[1pt]{\scriptsize\ttfamily refund, money back, \dots}};
\node[rep] (desc) at (-2.3,-1.75)
  {\textbf{Description}\\[1pt]{\scriptsize\itshape ``wants their money back''}};
\begin{scope}[on background layer]
  \node[draw=cInk!45, dashed, rounded corners=5pt, fit=(kp)(desc), inner sep=7pt] (tbox) {};
\end{scope}
\node[anchor=south, font=\footnotesize, text=cInk] at (tbox.north) {\textbf{Topic}~$t$ (two representations)};

\node[mat, draw=cRegex] (reg) at (2.5, 2.5)
  {\textbf{\textcolor{cRegex}{Regex}} {\scriptsize(baseline)}\\[1pt]{\scriptsize\ttfamily /refund/ /cancel/ \dots}};
\node[mat, draw=cEmbed] (emb) at (2.5, 0.0)
  {\textbf{\textcolor{cEmbed}{Embedding}}\\[1pt]{\scriptsize $\cos(\vec u,\vec t) > \tau$}};
\node[mat, draw=cLLM, line width=1pt] (llm) at (2.5,-2.5)
  {\textbf{\textcolor{cLLM}{LLM}}~$\star$\\[1pt]{\scriptsize ``Is $u$ about $t$?'' $\Rightarrow$ YES/NO}};

\node[io] (card) at (7.1,0)
  {\textbf{Coaching card}\\[2pt]{\scriptsize to the agent}};

\node[jdot] (j) at (-0.4,0) {};
\draw[ustream] (u.east) -- node[above=0.5pt, font=\scriptsize, text=cInk] {$u$} (j);
\draw[ustream] (j) to[out=70,in=190]  (reg.west);
\draw[ustream] (j) --                 (emb.west);
\draw[ustream] (j) to[out=-70,in=190] (llm.west);

\draw[edge] (kp.east)   to[out=0,in=170]   (reg.west);
\draw[edge] (kp.east)   to[out=0,in=150]   (emb.west);
\draw[edge] (kp.east)   to[out=-35,in=150] (llm.west);
\draw[edge] (desc.east) to[out=0,in=210]   (emb.west);
\draw[best] (desc.east) to[out=0,in=180]   (llm.west);

\draw[edge] (reg.east) to[out=-5,in=120]  (card.west);
\draw[edge] (emb.east) -- (card.west);
\draw[best] (llm.east) to[out=5,in=-120]  (card.west);
\end{tikzpicture}%
}
\caption{\textbf{Our evaluation space.} A live call yields a noisy utterance~$u$; a
predefined topic~$t$ is expressed two ways: a \textbf{keyphrase list} or a 
\textbf{natural-language description}. Every matcher scores the utterance against a topic representation: a
\textcolor{cRegex}{\textbf{regex}} baseline (keyphrases only), a zero-shot
\textcolor{cEmbed}{\textbf{embedding}} (cosine similarity), or an
\textcolor{cLLM}{\textbf{LLM}} prompted for a YES/NO decision. A positive match surfaces a
coaching card to the agent. We evaluate every applicable (utterance, representation) pairing;
the best~($\star$, blue) is an LLM reading a description.}
\label{fig:system}
\end{figure*}

Contact centers can use real-time ``agent-assist'' software to support human agents during a live call. Streaming Automatic Speech Recognition (ASR) transcribes the customer, and based on a curated list of topics, the system decides whether any of them is present in the conversation. When a topic matches the current utterance, the system displays a coaching card for the agent (Fig.~\ref{fig:system}). This requires low-latency processing of hundreds of topics simultaneously \citep{rawat2022caller,barrionuevo2026operators}. Additionally, the input text is a challenging ASR transcript of spontaneous speech containing disfluencies (e.g. ``my, my, my, the, my vehicle''), false starts, and residual recognition errors. These affect downstream processing pipelines \citep{ritter2011nertweets}, and the errors are non-negligible \citep{feng2021asrglue,shapira2025transcriptionnoise}. The standard approach is to clean the text, for example, with lexical normalization \citep{han2011lexnorm,vandergoot2021multilexnorm} or disfluency removal \citep{honnibal2014disfluency,zayats2016disfluency}.

We consider the \emph{binary topic--utterance matching} problem: given an utterance recognized with ASR, and a topic definition, decide whether the utterance matches the topic. To avoid additional latency, we process the raw ASR transcript. We compare two representations for the topic definition: the keyphrase lists that can be compiled into a regex matching program, and natural-language descriptions that can be processed by more advanced models. Both are evaluated on a human-annotated internal dataset. Various methods are benchmarked for matching utterances to topics: the regex baseline, zero-shot embedding similarity \citep{reimers2019sbert}, and Gemini LLM matchers \citep{geminiteam2025gemini25}. We summarize our empirical contributions:

\begin{enumerate}
\item \textbf{C1: to our knowledge, the first study of topic matching on real, noisy production
ASR, with the noise measured.} Previous resources are social-media text
\citep{baldwin2015wnut,derczynski2017wnutner,vandergoot2021multilexnorm}, read
or synthetic spoken-language understanding
\citep{bastianelli2020slurp,shon2022slue,feng2021asrglue}, or clean
crowdsourced intent sets
\citep{larson2019clinc,coucke2018snips,casanueva2020banking77}. None target the noisy textual input setting. We measure the noise directly (utterance length, $11.4\%$ disfluent repetitions, $24.2\%$ filler markers). However, our transcripts are sensitive customer data, so we release the evaluation protocol and findings rather than the data.

\item \textbf{C2: an efficient Flash-tier LLM is enough.} The LLM matchers
  beat the regex baseline (the best by $12.6$ $F_1$ points, $p<10^{-14}$) and
  every zero-shot encoder, even though encoders are the standard tool for
  semantic matching. The best two matchers are Flash-tier (Gemini-3-Flash and
  Gemini-2.5-Flash-Lite), and the heavier (e.g. Gemini-2.5-Pro) does not beat them. So a
  small, fast model is the practical choice for a
  latency-bounded deployment.

  \item \textbf{C3: the best way to represent and match a topic.} A topic can be
  written as a keyphrase list or as a natural-language description; we test every
  matcher against both and report the optimal combination. The strongest is an LLM
  reading a natural-language description ($F_1=0.847$), and the best representation
  differs by method---LLMs favor a description, embeddings a keyphrase list. How a
  topic is represented is thus as decisive as which matcher scores it.

\end{enumerate}
\section{Related Work}

\paragraph{Noisy user-generated text and normalization.}

Applying standard NLP to non-canonical text is a long-standing problem. Non-standard vocabulary and syntax break tools trained on edited corpora \citep{eisenstein2013badlanguage}, and off-the-shelf pipelines degrade sharply on tweets \citep{ritter2011nertweets}. The classic response is lexical normalization to a canonical vocabulary \citep{han2011lexnorm}, systems like MoNoise built on this idea \citep{vandergoot2017monoise}. Later work took normalization multilingual \citep{vandergoot2021multilexnorm}. For spontaneous \emph{speech}, the counterpart to orthographic normalization is disfluency detection and removal: finding repetitions, false starts, and self-corrections in transcribed speech \citep{honnibal2014disfluency,zayats2016disfluency}. These are the phenomena our noise statistics quantify. Our setting is spontaneous customer speech through call-center ASR, not social-media orthography. Rather than normalizing first, we match topics directly to achieve optimal latency performance, and leave a normalize-then-match comparison to future work.

\paragraph{ASR-error robustness and SLU.}

A parallel line studies how transcription errors propagate downstream. Spoken language understanding (SLU) benchmarks push toward natural speech \citep{bastianelli2020slurp,shon2022slue}. ASR-GLUE isolates natural language understanding (NLU) robustness under transcription noise \citep{feng2021asrglue}. Models trained on clean text degrade on ASR output, and recovering that performance takes deliberate effort \citep{cui2021asrrobustness,huang2020asrrobustembeddings,shapira2025transcriptionnoise}. We differ in using real contact-center ASR, not read, with binary topic matching as the downstream task.

\paragraph{Zero-shot classification and semantic matching.}

Zero-shot prompting \citep{brown2020gpt3} with instruction tuning \citep{wei2022flan} underpins the LLM-based paradigm, in which strong models proxy human judgment \citep{zheng2023judging,liu2023geval,gilardi2023chatgpt}. As a non-generative alternative we use zero-shot encoders \citep{reimers2019sbert}. Our set spans E5, BGE, MiniLM, and MPNet \citep{wang2022e5,xiao2024cpack,wang2020minilm,song2020mpnet}, chosen for their Massive Text Embedding Benchmark (MTEB) rankings \citep{muennighoff2023mteb}. To our best knowledge, no prior work systematically compares LLM matchers, zero-shot embeddings, and a regex baseline for semantic matching over noisy ASR.

\paragraph{Contact-center and intent-detection NLP.}

Our task is short-utterance intent detection. Canonical benchmarks define the family: CLINC150/OOS \citep{larson2019clinc}, Snips \citep{coucke2018snips}, Banking77 \citep{casanueva2020banking77}, and HWU64 \citep{liu2019benchmarking}, alongside short-text classification \citep{li2022textclass}. These use clean crowdsourced text. Closer to deployment, prior work classifies intent from call-center transcripts \citep{zhong2019callintent} and triggers agent-assist over live support ASR \citep{rawat2022caller,barrionuevo2026operators}. Related resources cover policy-driven dialogue \citep{chen2021abcd} and multi-annotator customer service \citep{feigenblat2021tweetsumm}. Our work combines these threads: an LLM matcher on noisy contact-center ASR for real-time agent assist, measured against a regex baseline and zero-shot embeddings.
\section{Task and Data}

\label{sec:task-data}

\subsection{Task formulation}
\label{sec:task}

We study \emph{binary topic--utterance matching} for real-time
contact center agent assist. During a live call, streaming ASR transcribes the customer, and for each predefined
topic the system decides whether the current utterance\footnote{By \emph{utterance} we mean a single finalized piece of
ASR output---one stretch of the customer's speech the recognizer has settled
on, rather than a whole speaker turn or the entire call. We match each one
on its own.}
is \emph{about} that topic. A positive decision surfaces a coaching card to
the agent. This runs concurrently for many topics per call and is
latency-sensitive \citep{rawat2022caller,barrionuevo2026operators}.
Formally, a matcher $f(u, d(t))\!\rightarrow\!\{0,1\}$ maps an utterance $u$
and a topic definition $d(t)$ to fire or no-fire; soft-output models produce
a score thresholded at $\tau$. We score against human gold as binary
classification, and because the classes are imbalanced
(Section~\ref{sec:data}) we foreground precision, recall, and $F_1$ over
accuracy. The
task is a form of short-utterance intent classification
\citep{li2022textclass,larson2019clinc,casanueva2020banking77,coucke2018snips},
but on unconstrained, noisy ASR.

\paragraph{Two topic representations.} A topic can be represented two ways.
The first is a \textbf{keyphrase list}: a curated set of trigger phrases
based on filler-tolerant regexes.
The second is a \textbf{natural-language description}: a short admin-authored
sentence of when the topic applies, scored YES/NO by a model
. This lets us ask which representation, paired with which
matcher, works best, and whether free-text descriptions can replace
hand-curated keyphrases.

\paragraph{Example.} For a topic like \emph{requesting a refund}, the
keyphrase list holds triggers such as \texttt{refund}, \texttt{money back},
and \texttt{reimburse}, while the natural language description reads ``The customer is asking
to be refunded or to get their money back.'' An utterance like ``so can I,
can I just get my money back for that?'' should match; ``my address is
forty-two oak street'' should not. (Examples are illustrative, not drawn from our real customer  data.)

\subsection{Data}
\label{sec:data}

\paragraph{Source.} Our benchmark comes from real English customer-service calls at
\textbf{11 companies} from various industries, covering \textbf{173 topics/coaching cards}. Across those
cards the keyphrase lists hold \textbf{3{,}658 keyphrases} (\textbf{2{,}447}
distinct surface forms), and the keyphrase regex runs over the full set
rather than a truncated sample.

\paragraph{Utterances are noisy ASR.} Inputs are ASR transcripts
of spontaneous customer speech. This is noisy user-generated text: disfluencies and repetitions, false
starts, residual recognition errors, and no reliable sentence structure.
These phenomena break off-the-shelf pipelines
\citep{ritter2011nertweets,han2011lexnorm} and propagate into downstream
understanding
\citep{bastianelli2020slurp,feng2021asrglue,shapira2025transcriptionnoise,cui2021asrrobustness}.

\paragraph{Noise characterization.} Aggregate statistics over the utterances
in the samples demonstrate the ``noisy'' nature of our data. The utterances
are short and variable: mean $15.8$ tokens, median $10$,
and $11.5\%$ have $\leq$3 tokens. An immediate word repetition (a stuttered
restart) appears in $11.4\%$, and $24.2\%$ contain a filler or discourse
marker (\emph{uh}, \emph{like}, \emph{you know}; $2.5\%$ of all tokens).
These statistics quantify spontaneous-speech \emph{disfluency}. True ASR
recognition errors are also present, our ASR system presents a word error rate (WER) of 14.26\% for English.

\subsection{Annotation and gold standard}
\label{sec:gold}

Every (utterance, topic) pair is labeled independently by three internal
annotators, who choose \emph{match}, \emph{no-match}, or \emph{ambiguous} (guidelines
in Appendix~\ref{sec:annotation}). The third option is deliberate: a short, disfluent
utterance is often genuinely underspecified, and forcing a binary choice would push
that uncertainty into the gold as noise. The gold label is the majority vote, and a
pair whose majority is \emph{ambiguous}, or that has no majority, is set aside rather
than guessed (73 pairs: 45 ambiguous, 28 without a majority). Of our
annotated pairs, \textbf{2{,}655} carry a decisive positive or negative gold
($\sim$24\% positive) and form the evaluation set, which indicates the imbalance that motivates
precision, recall, and $F_1$ over accuracy.

On the \textbf{2{,}655} fully voted pairs, inter-annotator
agreement is Fleiss' $\kappa=\textbf{0.660}$ across the three categories, which translates to ``substantial''
on the Landis--Koch scale \citep{fleiss1971kappa,landis1977agreement}. It sits below
perfect for a real reason: deciding whether a half-finished, disfluent utterance is
\emph{about} a topic is a judgment call that is hard for people, not only for models
\citep{derczynski2017wnutner}.

Together with authentic call center ASR, rather than the social media text \citep{baldwin2015wnut,vandergoot2021multilexnorm} or the read and synthetic
speech of SLU benchmarks \citep{bastianelli2020slurp,feng2021asrglue}, our
consensus-based, multi-annotator gold standard is what defines the new evaluation setting
we contribute.

\section{Methods}

We evaluate three matcher families on one shared benchmark, the same $2{,}655$-row human gold set from Section~\ref{sec:data}. The matcher families are (i) a regex baseline, (ii) zero-shot sentence-embedding encoders, and (iii) instruction-tuned large language model (LLM) matchers. Figure~\ref{fig:system} summarizes each family, the topic representation it uses, and how it decides.

\subsection{Regex baseline}
\label{sec:methods-regex}
Our baseline defines each topic by its keyphrase list and fires when an utterance matches any keyphrase under a word-boundary regex. The regex tolerates filler ``slop'' and normalizes hyphenation, punctuation, and numeric variants. We run it over the full keyphrase set, which contains $3{,}658$ keyphrases, $2{,}447$ distinct, across $173$ coaching cards, so the baseline reflects the keyphrase matcher as actually configured.

\subsection{Zero-shot sentence-embedding encoders}
\label{sec:methods-embed}
We embed the utterance and the topic, then score their cosine similarity, following the Sentence-BERT encoder tradition \citep{reimers2019sbert}. We use six widely adopted encoders zero-shot, with no fine-tuning: \texttt{e5-small/base-v2} \citep{wang2022e5}, \texttt{bge-small/base-en-v1.5} \citep{xiao2024cpack}, \texttt{all-MiniLM-L6-v2} \citep{wang2020minilm}, and \texttt{all-mpnet-base-v2} \citep{song2020mpnet}. We selected them by their standing on the Massive Text Embedding Benchmark (MTEB) \citep{muennighoff2023mteb}. We run two modes. The \emph{description} mode takes one cosine to the natural-language description. The \emph{keyphrase max-similarity} mode takes the max cosine over the topic's keyphrases. Each mode produces a continuous score that needs a threshold. Tuning that threshold on the test rows would bias the result, so we use $5$-fold stratified cross-validation (CV) \citep{kohavi1995crossval}. We select the max-$F_1$ threshold on four folds, apply it to the held-out fold, and aggregate. The stratification preserves the $\sim$24\% positive rate in each fold. 

\subsection{LLM matchers}
\label{sec:methods-llm}
We recast matching as a prompted YES/NO decision for LLMs. Instruction tuning makes this viable zero-shot \citep{brown2020gpt3,wei2022flan,gilardi2023chatgpt}. We evaluate four Gemini matchers: Gemini-2.5-Flash-Lite, Gemini-2.5-Flash, Gemini-2.5-Pro, and Gemini-3-Flash (preview) \citep{geminiteam2023gemini,geminiteam2024gemini15,geminiteam2025gemini25}---which together cover a range of speed and capability, the trade-off a real-time budget forces. All four use the same prompt (Appendix~\ref{sec:prompt}) at temperature $0$, and each emits a hard YES/NO that we use directly. A few responses come back unparseable ($\leq$9 per Gemini-2.5 matcher, none for Gemini-3), and we treat these as no-match (fail-closed). We hold the prompt, decoding, and parsing fixed across all four isolates the model as the source of any performance gap.

\subsection{Metrics}
\label{sec:methods-metrics}
The classes are imbalanced. Following work on skewed evaluation \citep{saito2015prplot,davis2006prroc,sokolova2009measures}, we center precision, recall, and $F_1$ over accuracy. We score hard-output methods at their native YES/NO decision and embeddings at the CV threshold. We add several agreement and significance measures: $95\%$ bootstrap confidence intervals (CIs) on $F_1$; McNemar tests for the headline contrasts; and $F_1$ on the fully adjudicated subset.

\subsection{Cost and latency}
\label{sec:methods-cost}
For each matcher we measure per-decision latency and cost on $200$ topic--utterance pairs sampled from the benchmark, discarding a short warm-up. Latency is the end-to-end time for one $(\text{utterance},\text{topic})$ decision under identical conditions: for the LLM matchers, the round trip to the Google Gemini API (temperature $0$, one \texttt{YES}/\texttt{NO} token); for embeddings, the utterance encode; for regex, the pattern match. We report it as a relative comparison, not a production service-level objective. We report the marginal API cost per $1{,}000$ decisions for the LLM matchers, from the published per-token prices\footnote{Gemini API price list, accessed 2026-08-27.} applied to the measured input and output tokens (output includes billed thinking tokens). The underlying compute cost of serving any matcher (instance, CPU/GPU, memory) varies by deployment and is out of scope; we assume a comparable instance-serving cost across all approaches and report only the differential API cost, which the LLM matchers alone incur. The figures are therefore the additional per-decision API cost on top of a common serving baseline, not a total cost of ownership.

\section{Results and Discussion}

\label{sec:results}

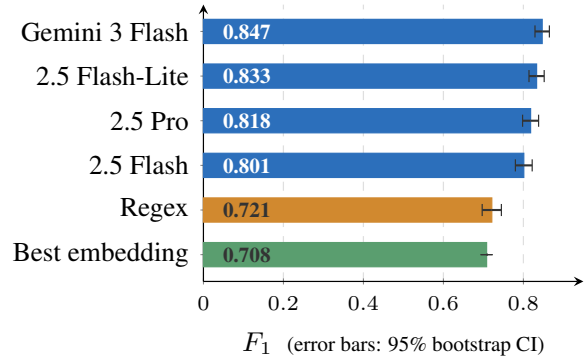
\begin{figure}[t]
\centering
\resizebox{\columnwidth}{!}{%
\begin{tikzpicture}
\begin{axis}[
  width=6cm, height=4.8cm,
  xbar, /pgf/bar shift=0pt, bar width=8pt,
  xmin=0, xmax=0.95,
  ymin=0.4, ymax=6.6,
  ytick={1,2,3,4,5,6},
  yticklabels={Best embedding, Regex, 2.5 Flash,
               2.5 Pro, 2.5 Flash-Lite, Gemini 3 Flash},
  yticklabel style={font=\footnotesize},
  xlabel={$F_1$ \;\scriptsize(error bars: 95\% bootstrap CI)},
  xlabel style={font=\footnotesize},
  xtick={0,0.2,0.4,0.6,0.8},
  xticklabel style={font=\scriptsize},
  xmajorgrids=true, grid style={dashed, gray!25},
  tick align=outside, major tick length=2pt,
  axis lines=left, clip=false,
  error bars/x dir=both, error bars/x explicit,
  error bars/error bar style={line width=0.7pt, cInk},
]
\addplot+[draw=cLLM, fill=cLLM, mark=none] coordinates {
  (0.801,3) +- (0.021,0)
  (0.818,4) +- (0.020,0)
  (0.833,5) +- (0.019,0)
  (0.847,6) +- (0.018,0)
};
\addplot+[draw=cRegex, fill=cRegex, mark=none] coordinates {
  (0.721,2) +- (0.024,0)
};
\addplot+[draw=cEmbed, fill=cEmbed, mark=none] coordinates {
  (0.708,1) +- (0,0)
};
\node[anchor=west, font=\scriptsize\bfseries, text=white] at (axis cs:0.02,6) {0.847};
\node[anchor=west, font=\scriptsize\bfseries, text=white] at (axis cs:0.02,5) {0.833};
\node[anchor=west, font=\scriptsize\bfseries, text=white] at (axis cs:0.02,4) {0.818};
\node[anchor=west, font=\scriptsize\bfseries, text=white] at (axis cs:0.02,3) {0.801};
\node[anchor=west, font=\scriptsize\bfseries, text=cInk]  at (axis cs:0.02,2) {0.721};
\node[anchor=west, font=\scriptsize\bfseries, text=cInk]  at (axis cs:0.02,1) {0.708};
\end{axis}
\end{tikzpicture}%
}
\caption{Main results: $F_1$ on the human-gold benchmark. Error bars are
$95\%$ bootstrap CIs for the native-decision methods;
the \textcolor{cEmbed}{\textbf{best embedding}} bar is a 5-fold cross-validated point
estimate (fold $\sigma\!=\!0.027$; Table~\ref{tab:keyphrases}) and is shown without a
bootstrap bar as it is not a native-decision method. \textcolor{cLLM}{\textbf{LLM matchers}}
(blue) lead the \textcolor{cRegex}{\textbf{regex}} baseline (orange) and the best
of six zero-shot embedding encoders (green); the two lightweight Flash-tier matchers top the
board and the LLM--regex gap is significant ($p<10^{-6}$ for all four matchers).}
\label{fig:results}
\end{figure}

Figure~\ref{fig:results} gives the headline: the best $F_1$ each matcher reaches. Tables~\ref{tab:keyphrases} and~\ref{tab:desc} break the picture out by topic representation, so every matcher--representation pairing is visible. Because the classes are imbalanced ($\sim$24\% positive), we report precision, recall, and $F_1$; embeddings are scored at a cross-validated threshold and the other matchers at their native YES/NO decision, and unparseable LLM replies (at most nine per model) count as no-match.

\subsection{An LLM on a description wins}
The strongest matcher is an LLM reading a natural-language description. Gemini-3-Flash reaches $F_1=0.847$, ahead of the regex baseline ($0.721$) by $12.6$ points and of the best embedding ($0.708$). The margin over regex is large and highly significant (exact McNemar $p<10^{-6}$ for every LLM, and $p<10^{-14}$ for the two best), and it is almost all recall: the LLM catches paraphrases and disfluent phrasings that a fixed keyphrase list misses. Surprisingly, embeddings come last though being the usual backbone of semantic matching in many other tasks.

\begin{table}[t]
\centering
\small
\begin{tabular}{@{}lccc@{}}
\toprule
Keyphrases & P & R & $F_1$ \\
\midrule
Regex (baseline)  & 0.850 & 0.625 & 0.721 \\
\midrule
all-MiniLM-L6-v2  & 0.689 & 0.731 & 0.708 \\
e5-base-v2        & 0.673 & 0.743 & 0.706 \\
bge-base-en-v1.5  & 0.669 & 0.748 & 0.706 \\
bge-small-en-v1.5 & 0.657 & 0.762 & 0.705 \\
mpnet-base-v2     & 0.750 & 0.637 & 0.689 \\
e5-small-v2       & 0.620 & 0.746 & 0.677 \\
\midrule
Gemini 3 Flash    & 0.839 & 0.752 & \textbf{0.793} \\
2.5 Pro           & 0.845 & 0.725 & 0.780 \\
2.5 Flash-Lite    & 0.882 & 0.638 & 0.740 \\
2.5 Flash         & 0.898 & 0.603 & 0.721 \\
\bottomrule
\end{tabular}
\caption{All matchers on the \textbf{keyphrase} representation. Embeddings are 5-fold cross-validated; regex and LLMs use their native decision. Best $F_1$ in bold.}
\label{tab:keyphrases}
\end{table}

\begin{table}[t]
\centering
\small
\begin{tabular}{@{}lccc@{}}
\toprule
Natural-language description & P & R & $F_1$ \\
\midrule
e5-base-v2        & 0.687 & 0.635 & 0.660 \\
all-MiniLM-L6-v2  & 0.660 & 0.640 & 0.649 \\
bge-small-en-v1.5 & 0.646 & 0.640 & 0.641 \\
bge-base-en-v1.5  & 0.615 & 0.657 & 0.635 \\
e5-small-v2       & 0.646 & 0.616 & 0.627 \\
mpnet-base-v2     & 0.604 & 0.626 & 0.613 \\
\midrule
Gemini 3 Flash    & 0.866 & 0.829 & \textbf{0.847} \\
2.5 Flash-Lite    & 0.887 & 0.786 & 0.833 \\
2.5 Pro           & 0.827 & 0.809 & 0.818 \\
2.5 Flash         & 0.931 & 0.703 & 0.801 \\
\bottomrule
\end{tabular}
\caption{All matchers on the \textbf{natural-language description} representation. Regex is omitted as it matches keyphrase lists only. Embeddings are 5-fold cross-validated. Best $F_1$ in bold.}
\label{tab:desc}
\end{table}

\paragraph{Representation matters as much as the matcher.}
Tables~\ref{tab:keyphrases} and~\ref{tab:desc} show that the best representation is not the same for every method. LLMs are strongest on the description (Gemini-3-Flash $0.847$ vs.\ $0.793$ on keyphrases; Gemini-2.5-Flash-Lite $0.833$ vs.\ $0.740$); embeddings are strongest on keyphrases ($0.708$ vs.\ $0.660$); regex applies only to keyphrases, since a description is not a list of patterns. The optimal pairing is therefore an LLM with a natural language description, and how a topic is represented is as consequential as which matcher scores it.

\paragraph{A lighter model is enough.}
The two best systems are the two smallest LLMs, Gemini-3-Flash and Gemini-2.5-Flash-Lite, and no larger model beats them. Gemini-2.5-Flash-Lite is even nominally ahead of the heavier Gemini-2.5-Pro and Gemini-2.5-Flash ($p=0.014$/$0.019$). This interesting finding suggests that a small Flash-tier model is enough for this task while heavier, most costly LLMs offer no obvious benefits.

\paragraph{Cost and latency.}
Table~\ref{tab:latency-cost} prices the ``lighter is enough'' finding. Gemini-2.5-Pro is Pareto-dominated: its mandatory thinking (${\sim}525$ output tokens on a \texttt{YES}/\texttt{NO} question) makes it the slowest (p50 $5.6$\,s) and by far the costliest (\$$5.38$ per $1{,}000$ decisions), yet it does not lead on $F_1$. Gemini-2.5-Flash-Lite sits at the opposite corner: p50 $0.35$\,s and \$$0.010$ per $1{,}000$, i.e.\ ${\sim}16\times$ faster and ${\sim}500\times$ cheaper than 2.5-Pro, at $F_1$ within $0.02$ of the best matcher. Embeddings and regex are faster still and self-hosted, but trail on $F_1$ (Tables~\ref{tab:keyphrases},~\ref{tab:desc}). For a latency- and cost-bounded real-time deployment, a Flash-tier LLM on a description is the practical operating point.

\begin{table}[t]
\centering\small
\setlength{\tabcolsep}{4.5pt}
\begin{tabular}{@{}lrrrrr@{}}
\toprule
Matcher & \multicolumn{2}{c}{Latency (ms)} & \multicolumn{2}{c}{Tokens} & \$/1k \\
\cmidrule(lr){2-3}\cmidrule(lr){4-5}
 & p50 & p95 & in & out & \\
\midrule
Gemini 3 Flash & 1742 & 5018 & 100 & 165 & 0.55 \\
2.5 Flash-Lite & 346 & 528 & 100 & 1 & 0.010 \\
2.5 Pro & 5580 & 8365 & 100 & 525 & 5.38 \\
2.5 Flash & 461 & 771 & 100 & 1 & 0.033 \\
e5-base-v2 & 29 & 32 & -- & -- & -- \\
all-MiniLM-L6-v2 & 6 & 7 & -- & -- & -- \\
Regex & 0 & 0 & -- & -- & -- \\
\bottomrule
\end{tabular}
\caption{Per-decision latency and API cost on 200 sampled topic--utterance pairs. LLMs score the natural-language description; embedding latency is the per-utterance encode and regex the keyphrase match. Gemini \$/1k is the marginal API cost (prices accessed 2026-08-27; output includes billed thinking tokens). Embeddings and regex are self-hosted and incur no API cost. Latency is measured under identical conditions from one client to the Google Gemini API; embeddings and regex are timed locally on a single CPU core.}
\label{tab:latency-cost}
\end{table}

\subsection{Across three matchers}
Our central finding is that LLM matchers score significantly higher on this noisy benchmark. What is clear mechanistically is that regex matches only literal keyphrases and cannot absorb disfluencies or recognition errors, and that cosine similarity rewards surface overlap rather than topic semantics when processing noisy textual input as in our setting. 

\section{Conclusion}

We studied topic--utterance matching on noisy call-center ASR transcripts, comparing a regex baseline, zero-shot embeddings, and Gemini LLM matchers across two ways of describing a topic. The clear winner is an LLM reading a natural-language description: Gemini-3-Flash reaches an $F_1$ of $0.847$, well above the regex baseline ($0.721$) and the best embedding ($\approx0.71$). Our comprehensive benchmark demonstrates that lightweight LLMs are enough for this task. The smallest Flash-tier matchers lead, while larger ones do not present superior performance. Additionally, we prove that the topic representation matters as much as the matcher: LLMs are strongest with a written description, while embeddings work better with a keyphrase list. Because the calls are sensitive customer data, we present this evaluation protocol and findings rather than the data itself. We hope this work provides insights to other practitioners working under a similar setting.

\section*{Limitations}

\paragraph{Scope.} All data are English and from a single domain, contact-center agent assist (11 companies, 173 coaching cards); we make no cross-lingual or cross-domain claims. This is an accuracy result: without a controlled clean-versus-noisy comparison, we do not isolate noise robustness from general semantic capability.

\paragraph{Models.} Our best matchers are proprietary Gemini API models \citep{geminiteam2025gemini25}, and the very best (Gemini-3-Flash) is a preview model whose numbers are a snapshot. Our conclusions do not hinge on it: the generally available Gemini-2.5-Flash-Lite already beats every non-LLM baseline, and we anchor the comparison to a fully reproducible regex baseline and open-weight embeddings \citep{reimers2019sbert}. We do not test other LLM API services beyond Gemini due to our data safety policy.

\paragraph{Data availability.} The underlying transcripts contain sensitive customer information and cannot be released. To support reproducibility within this constraint, we release the full evaluation protocol, prompts, matcher configurations, and aggregate statistics, and we report results only in aggregate; access to the data may be considered on a controlled basis subject to the service's data-processing agreements.

\section*{Ethics and Broader Impact}

The data is derived from customer--agent dialogues about sensitive topics and processed in accordance with the service's data-processing and consent procedures; we use only the transcript text and do not seek to re-identify any customers or agents. We report only aggregated results, model performance scores, and evaluation methodology, not the data itself. Gold labels were obtained from internal specialized annotators who were compensated fairly.

\bibliography{custom}

\appendix

\section{Evaluation Prompt}
\label{sec:prompt}
All LLM matchers receive the identical prompt below (temperature $0$). The
system instruction fixes the task and the strict \texttt{YES}/\texttt{NO}
output format; the user template is filled per example with the topic
description and the customer utterance.

\paragraph{System instruction.}
\begin{quote}\ttfamily\small
You are a precise judge evaluating whether a customer utterance matches a
topic description. The topic description defines what a customer might say
when they are discussing that topic. You must reply with exactly one word:
YES or NO. Do not add any explanation.
\end{quote}

\paragraph{User template.}
\begin{quote}\ttfamily\small
Topic description:\\
\{description\}\\[2pt]
Customer utterance:\\
``\{utterance\}''\\[2pt]
Does the utterance match the topic? Reply YES or NO.
\end{quote}

\section{Annotation Guidelines}
\label{sec:annotation}
Each item shows one customer utterance and one topic (its natural-language
description). Up to three annotators independently answer a single
question, presented verbatim as:
\begin{quote}\itshape\small
Does the utterance fit the description? Would you say the utterance fits
under the topic described?
\end{quote}
with response options \textbf{yes} (positive / match), \textbf{no}
(negative / no-match), or \textbf{ambiguous} when the utterance is too
underspecified to decide, plus an optional free-text notes field for
rationale. Annotators judge \emph{semantic} fit---whether the customer is
talking about the topic---rather than surface keyword overlap, and are
instructed to rely only on the utterance itself (no surrounding call
context). The per-item gold label is the majority vote over the available
annotators. Rows whose majority label is \textbf{ambiguous}, or with no
majority (\textbf{disputed}), are excluded from the usable evaluation set;
rows with only a single annotator so far are used as per-row gold but
flagged and excluded from all inter-annotator agreement statistics
(Section~\ref{sec:data}).

\end{document}